# Marvin: A Heuristic Search Planner with Online Macro-Action Learning

**Andrew Coles**          ANDREW.COLES@CIS.STRATH.AC.UK
**Amanda Smith**         AMANDA.SMITH@CIS.STRATH.AC.UK
*Department of Computer and Information Sciences,*
*University of Strathclyde,*
*26 Richmond Street, Glasgow, G1 1XH, UK*

## Abstract

This paper describes Marvin, a planner that competed in the Fourth International Planning Competition (IPC 4). Marvin uses action-sequence-memoisation techniques to generate macro-actions, which are then used during search for a solution plan. We provide an overview of its architecture and search behaviour, detailing the algorithms used. We also empirically demonstrate the effectiveness of its features in various planning domains; in particular, the effects on performance due to the use of macro-actions, the novel features of its search behaviour, and the native support of ADL and Derived Predicates.

## 1. Introduction

One of the currently most successful approaches to domain-independent planning is forward-chaining heuristic search through the problem state space. Search is guided by a heuristic function based on an appropriate relaxation of the planning problem. Different relaxations have been explored (Bonet & Geffner, 2000; McDermott, 1996; Hoffmann & Nebel, 2001; Helmert, 2004) and have been shown to result in more or less informative heuristic functions. A common relaxation is to ignore the delete lists of actions in the problem domain, resulting in an abstracted problem domain comprised of *relaxed actions*. A given state can then be evaluated by counting the number of relaxed actions needed to reach the goal state from the given state. Hoffmann and Nebel (2001) present a search strategy called Enforced Hill-Climbing (EHC) which, coupled with a relaxation of this kind, has been proven empirically to be an effective strategy in many planning domains. Their planner, FF, performed with great success in the Second and Third International Planning Competitions (Bacchus, 2001; Long & Fox, 2003). In this paper we present our planner, Marvin, which builds upon this search approach.

The EHC search strategy performs gradient-descent local search, using breadth-first search to find action sequences leading to strictly-better states should no single-action step be able to reach one. This embedded exhaustive-search step is one of the bottlenecks in planning with this approach. We present an approach that, through memoising the plateau-escaping action sequences discovered during search, can form macro-actions which can be applied later when plateaux are once-again encountered. In doing so, the planner can escape from similar plateaux encountered later, without expensive exhaustive search. The resulting planner is called Marvin.

We begin this paper with a brief review of FF's search behaviour to provide the background for our approach. We then introduce the main features of Marvin, explaining how its search behaviour differs from that of FF. We describe the three main contributions made by Marvin, detailing the key algorithms and their effects on performance. Marvin can plan in STRIPS and ADL domains, and it can also handle the derived predicates of PDDL2.2. We describe the way in which domains containing derived predicates and ADL are handled without first being reduced to STRIPS domains. Finally, we discuss the results obtained by Marvin in the Fourth International Planning Competition (IPC 4) (Hoffmann & Edelkamp, 2005), and provide additional ablation studies to assess the impact of its various features on planning performance across a selection of domains.

## 2. Background

In this section, we give an overview of the background for this work. First, forward-chaining heuristic planning is defined, and existing work in this area described; with particular attention paid to the planner FF. This is followed by an introduction to macro-actions.

### 2.1 Forward-Chaining Planning

Formally, forward-chaining planning can be described as search through a landscape where each node is defined by a tuple $< S, P >$. $S$ is a world state comprised of predicate facts and $P$ is the plan (a series of ordered actions) used to reach $S$ from the initial state. Search begins from the initial problem state, corresponding to a tuple $< S_0, \{\} >$.

Edges between pairs of nodes in the search landscape correspond to applying actions to lead from one state to another. When an action $A$ is applied to a search space node $< S, P >$ the node $< S', P' >$ is reached, where $S'$ is the result of applying the action $A$ in the state $S$ and $P'$ is determined by appending the action $A$ to $P$. Forward-chaining search through this landscape is restricted to only considering moves in a forwards direction: transitions are only ever made from a node with plan $P$ to nodes with a plan $P'$ where $P'$ can be determined by adding (or 'chaining') actions to the end of $P$.

As unguided search in this manner is prohibitively expensive in all but the smallest problems, heuristics are used to guide search. Commonly, a heuristic value is used to provide a goal distance estimate from a node $< S, P >$ to a node $< S', P' >$ in which $S'$ is a goal state.

### 2.2 Heuristics for Forward-Chaining Planning

Many of the heuristics used to guide forward-chaining planners are based around solving an abstraction of the original, hard, planning problem with which the planner is presented. The most widely used abstraction involves planning using 'relaxed actions', where the delete effects of the original actions are ignored. FF, HSP (Bonet & Geffner, 2000) and UnPOP (McDermott, 1996) use relaxed actions as the basis for their heuristic estimates, although FF was the first to count the number of relaxed actions in a relaxed plan connecting the goal to the initial state. Although ignoring delete lists turns out to be a powerful relaxation, at least for a class of planning domains, other relaxations are possible. More recently, work has been done using an abstraction based on causal graph analysis (Helmert, 2004).

The initial approaches for calculating goal distance estimates, taken by planners such as HSP, calculated the cost of reaching the goal state from the state to be evaluated by doing a forwards reachability analysis from the state until the given goal appears. Two heuristics can be derived from this information: either the maximum of the steps-to-goal values—an admissible heuristic; or the sum of the steps-to-goal values—an inadmissible heuristic, which in practice is more informative.

```
 1: Procedure: EHCSearch
 2: open_list = [initial_state];
 3: best_heuristic = heuristic value of initial_state;
 4: while open_list not empty do
 5:     current_state = pop state from head of open_list;
 6:     successors = the list of states visible from current_state;
 7:     while successors is not empty do
 8:         next_state = remove a state from successors;
 9:         h = heuristic value of next_state;
10:         if next_state is a goal state then
11:             return next_state;
12:         end if
13:         if h better than best_heuristic then
14:             clear successors;
15:             clear open_list;
16:             best_heuristic = h;
17:         end if
18:         place next_state at back of open_list;
19:     end while
20: end while
```

Figure 1: Enforced Hill-Climbing Search

The disadvantage of the latter of these approaches is that it ignores any positive interactions (shared actions) between the action sequences for each goal: it is this problem which was addressed by the heuristic used in FF. In FF, a planning graph (Blum & Furst, 1995) is built forward from the current state using relaxed actions—this is known as a relaxed planning-graph (RPG). A relaxed plan (one using relaxed actions) to achieve the goal state can be extracted from the RPG in polynomial time; the number of actions in this plan can be used to provide the heuristic value. As Graphplan does not provide a guarantee that the plan found will contain the optimum number of sequentialised actions (only that it will have the optimum makespan) the heuristic returned is inadmissible, but in practice the heuristic is more informative than any of those used previously.

### 2.3 Enforced Hill Climbing Search

Along with a heuristic based on relaxed planning graphs, FF introduced the Enforced Hill Climbing (EHC) algorithm, illustrated in Figure 1. EHC is based on the commonly used hill-climbing algorithm for local search, but differs in that breadth-first search forwards from the global optimum is used to find a sequence of actions leading to a heuristically better successor if none is present in the immediate neighbourhood.

The key bottleneck in using EHC is where the search heuristic cannot provide sufficient guidance to escape a plateau[1] in a single action step, and breadth-first search is used until a suitable action sequence is found. Characteristically, EHC search consists of prolonged periods of exhaustive search, bridged by relatively quick periods of heuristic descent.

---

1. In this work, a plateau is defined to be a region in the search space where the heuristic values of all successors is greater than or equal to the best seen so far.

In practice, EHC guided by the RPG heuristic is an effective search strategy in a number of domains. Work has been done (Hoffmann, 2005) on analysing the topology of the local-search landscape to investigate why it is an effective heuristic, as well as identifying situations in which it is weak.

## 2.4 Exploiting the Structure of a Relaxed Plan

The actions in the relaxed plan to the goal from a given state can be used to provide further search advice. YAHSP (Vidal, 2004), a planner that produced interesting results in the Fourth International Planning Competition (IPC 4), makes use of the actions of the relaxed plan to suggest actions to add to the current plan to reach the goal. In FF, the notion of 'helpful actions' is defined—those that add a fact added by an action chosen at the first time unit in the relaxed plan. In each state encountered during search, a number of actions are applicable, some of which are irrelevant; i.e. they make no progress towards the goal. By only considering the helpful actions when determining the successors to each state, when performing EHC, the number of successor states to be evaluated will be reduced.

Restricting the choice of actions to apply only to those that are 'helpful' further reduces the completeness of EHC, beyond what would be the case if all applicable actions were considered. In practice it is observed, however, that the cases where EHC using only helpful actions is unable to find a plan correlate with the cases where EHC with all the applicable actions would be unable to find a plan.

## 2.5 Guaranteeing Completeness in FF

FF first attempts to search for a solution plan by performing Enforced Hill-Climbing (EHC) search from the initial state towards the goal state. As discussed earlier, EHC uses hill-climbing local search guided by the RPG heuristic while a strictly-better successor can be found. As soon as no strictly better successor can be found, FF has entered a plateau, and breadth-first search is used until an improving state is found. In directed search spaces, EHC can lead the search process in the wrong direction and to dead-ends; i.e. the open list is empty, but no goal state has been found. In these cases FF resorts to best-first search from the initial state, thereby preserving completeness.

## 2.6 Macro-Actions in Planning

A macro-action, as used in planning, is a meta-action built from a sequence of action steps. In forward-chaining planning, applying a macro-action to a state produces a successor corresponding to the application of a series of actions. In this way, the use of macro-actions can be thought of as extending the neighbourhood of successors visible from each state to selectively introduce states which hitherto would only have been visible after the application of several steps. If the additional states introduced are chosen effectively, an increase in planner performance can be realised; whereas if the additional states are chosen poorly, the performance of the planner decreases due to the increased branching factor.

The use of macro-actions in planning has been widely explored. Most techniques use an off-line learning approach to generate and filter macro-actions before using them in search. Early work on macro-actions began with a version of the STRIPS planner—known as ABSTRIPS (Fikes & Nilsson, 1971)—which used previous solution plans (and segments thereof) as macro-actions in solving subsequent problems. MORRIS (Minton, 1985) later extended this approach by adding some filter-ing heuristics to prune the generated set of macro-actions. Two distinct types of macro-actions were identified in this approach: S-macros—those that occur frequently during search—and T-macros— those that occur less often but model some weakness in the heuristic. Minton observed that the T-macros, although used less frequently, offered a greater improvement in search performance. The REFLECT system (Dawson & Siklossy, 1977) took the alternative approach of forming macro-actions based on preprocessing of the domain. All sound pairwise combinations of actions were considered as macro-actions and filtered through some basic pruning rules. Due to the small size of the domains with which the planner was reasoning, the number of macro-actions remaining following this process was sufficiently small to use in planning.

More recent work on macro-actions includes that on Macro-FF (Botea, Enzenberger, Muller, & Schaeffer, 2005). Macro-actions are extracted in two ways: from solution plans; and by the identification of statically connected abstract components. An offline filtering technique is used to prune the list of macro-actions. Another recent approach to macro-action generation (Newton, Levine, & Fox, 2005) uses a genetic algorithm to generate a collection of macro-actions, and then filters this collection through an offline filtering technique similar to that used by Macro-FF.

## 3. Marvin's Search Behaviour

Marvin's underlying search algorithm is based on that used by FF: forward-chaining heuristic search using the RPG heuristic. However, Marvin includes some important modifications to the basic FF algorithm. These are: a *least-bad-first* search strategy for exploring plateaux, a *greedy best-first* strategy for searching when EHC fails and the development and use of *plateau-escaping macro-actions*.

As in FF the first approach to finding a solution plan is to perform EHC search using only helpful actions. The first successor with a heuristic strictly better than the best so far is taken, should one be found. If one is not found, then a plateau has been encountered, and a form of best-first search using helpful actions is used (instead of the breadth-first search of FF) to try to find an action sequence to escape from it. Because the states on a plateau can never improve on the heuristic value of the node at the root of the plateau, we call this *least-bad-first* search.

If the EHC approach is unable to find a plan, Marvin resorts to a *modified* form of best-first search using all the actions applicable in each state. This expands the first strictly better successor whilst keeping the current state for further expansion later if necessary. We call this strategy *greedy best-first search*. As can be seen in the graphs in Section 6.2, in some of the IPC 4 domains our modifications enable the solution of problems that are unsolvable for best-first search.

During the EHC search strategy, Marvin uses *plateau-escaping macro-actions* learned from previous searches of similar plateaux. These can be applied in the same way as atomic actions to traverse plateaux in a single step. Plateau-escaping macro-actions are learned online and the planner must decide which ones are likely to be applicable at which points during search. In Section 6.5 we show that plateau-escaping actions can yield performance benefits. Their power depends on the structure of the search space and the ability of the planner to learn re-usable macro-actions.

Least-bad-first search on plateaux, greedy best-first search and plateau-escaping macro-actions are the three main features of Marvin distinguishing its basic search strategy from that of other forward heuristic search-based planners. We now discuss these three features in more detail before going on to describe how they can be exploited in the context of ADL domains and domains involving derived predicates.

Figure 2: Least-bad-first search versus breadth-first search on a plateau. Black nodes are those expanded by breadth-first search. Dotted blue/grey nodes are those expanded by both breadth-first and least-bad-first search. Solid blue/grey nodes are those expanded by only least-bad-first search. It can be seen that least-bad-first search leads to a better exit node than does breadth-first search.

## 3.1 Least-Bad-First Search on Plateaux

A plateau is encountered when all of the successor nodes of a given current node have a heuristic value that is the same as, or worse than, that of the current node. The notion of *best* in this context relates to the successor with the heuristic value closest to that of the parent state. This is called least-bad-first search because no chosen successor can make heuristic progress, but some choices are less negative than others. The successor chosen in least-bad-first search will have least negative impact on the current state and therefore is more likely to be on the best path to the goal. When breadth-first search is used, the exit state that is reached might be further from the goal than the exit state reached when the state with the least negative impact is always expanded next.

In Figure 2 we show the order in which states are explored using least-bad-first search relative to breadth-first search. It can be observed that, using least-bad-first search, the exit state reached has a better heuristic value than that reached using the breadth-first search in FF. It can be expected that this sometimes leads to better quality plans being found. Our results in Section 6.3 show that, indeed, using least-bad-first search we sometimes find shorter plans than FF finds using its standard breadth-first strategy on plateaux.

## 3.2 Greedy Best-First Search when EHC Fails

As in FF, Marvin resorts to best-first search if EHC proves unable to find a solution. This approach maintains the completeness of the planner in the cases where the use of EHC with helpful actions would otherwise render the search incomplete. Two other planners in IPC 4 used variations on the best-first search algorithm, YAHSP (Vidal, 2004) and Fast-Downward (Helmert, 2006). Unlike Marvin, however, in these two planners there is no incomplete search step (such as EHC) before using a modified best-first search algorithm. In YAHSP, conventional WA* search is used but within the search queue, states reached by applying a helpful action in their parent state are ordered before those which were not. In Fast-Downward, a 'deferred heuristic evaluation' strategy is used, where states are inserted into the search queue with the heuristic value of their parent state; the actual heuristic cost of the state is then only calculated when the state is expanded.

In Marvin the best-first strategy is modified by greedily expanding the first successor found with a better heuristic than its parent state, but retaining the parent so that its remaining children can be evaluated later if necessary. The effect of this is similar to the approach taken in Fast-Downward, and would lead to the nodes in the search space being visited in the same order. The approach taken in Marvin, however, allows a smaller search queue to be maintained, as nodes are not necessarily inserted into the search queue for each successor node reachable from a state.

Whenever a successor state is generated and evaluated (by calculating its heuristic value), one of two things happens:

- If the successor has a heuristic better than its parent, the successor is placed at the front of the search queue, with its parent state behind it (along with a counter variable, noting how many successors have already been evaluated); and the search loop is then restarted from the successor state.

- If the successor has a heuristic no better than its parent, the successor is inserted into the search queue in its appropriate place (stable priority-queue insertion, ordered by heuristic value). The process then carries on evaluating the successors of the parent state.

The pseudo-code for this can be seen in Figure 3. The approach is inspired by the idea of taking the first strictly-better successor when performing EHC search, with the benefit that the number of heuristic evaluations to be performed is potentially reduced by considering fewer successors to each state. It differs from EHC in that, to maintain completeness, the parent state is not discarded—it is placed back in the queue to have its other successors evaluated later if necessary. Theoretically, if EHC search on a given problem does not encounter any plateaux, and any pruning from selecting only the helpful actions is ignored, then using greedy best-first search on that problem would visit the same number of nodes and evaluate the same number of successors. If a plateau was encountered, however, the search behaviour would differ as EHC would only consider states reachable from the state at the start of the plateau.

Another effect of the greedy best-first search is that the search focusses on exploring in a given direction. As has been described, as soon as a successor node is found with a heuristic value better than that of its parent, then the further expansion of the parent node is postponed and the successor node is expanded. The practical consequence of this is that as the search queue does not contain the other equally good successors, any search forward from a successor state will not be sidetracked by also having to search forward from its sibling states. The parent node will be re-visited, and the other sibling nodes added, but only if it proves heuristically wise to do so—that is, if searching forward from the successor node is not making heuristic progress.

### 3.3 Plateau-Escaping Macro-Actions

Due to the nature of the relaxed problem used to generate the RPG heuristic there are aspects of the original problem that are not captured. Thus, when the RPG heuristic is used to perform EHC, plateaux are often encountered. On plateaux, the RPG heuristic value of all successor states is the same as, or worse than, the heuristic value of the current state. The nature of the plateaux encountered, and whether EHC is able to find a path to escape from them, is influenced by the properties of the planning domain (Hoffmann, 2001).

1:  **Procedure: GreedyBFS**
2:  insert (state=initial_state, h=initial_heuristic, counter=0) into search_queue;
3:
4:  **while** search_queue not empty **do**
5:      current_queue_entry = pop item from front of search_queue;
6:      current_state = state from current_queue_entry;
7:      current_heuristic = heuristic from current_queue_entry;
8:      starting_counter = counter from current_queue_entry;
9:      applicable_actions = array of actions applicable in current_state;
10:
11:     **for all** index ?i in applicable_actions $\geq$ starting_counter **do**
12:         current_action = applicable_actions[?i];
13:         successor_state = current_state.apply(current_action);
14:
15:         **if** successor_state is goal **then**
16:             return plan and exit;
17:         **end if**
18:         successor_heuristic = heuristic value of successor_state;
19:
20:         **if** successor_heuristic < current_heuristic **then**
21:             insert (current_state, current_heuristic, ?i + 1) at front of search_queue;
22:             insert (successor_state, successor_heuristic, 0) at front of search_queue;
23:             break for;
24:
25:         **else**
26:             insert (successor_state, successor_heuristic, 0) into search_queue;
27:         **end if**
28:     **end for**
29: **end while**
30: exit - no plan found;

Figure 3: Pseudo-code for greedy best-first search

Ignoring the delete effects of the `pickup` action in the Gripper domain creates a problem in which, in a given state, it is possible to pick up many balls using one gripper, so long as the gripper is initially available: the delete effect of the action, marking the gripper as no longer available, is removed. The relaxed plan in the initial problem state is `pickup` all the balls with one gripper, `move` to the next room, then `drop` them all. The length of this plan, the heuristic value of the initial state, is $n + 1 + n$, that is $2n + 1$ (where $n$ is the number of balls). If, in the initial state, a ball is picked up using one of the grippers, the relaxed plan for the resulting state will be to `pickup` the remaining balls in the other gripper, `move` to the second room and then `drop` them all; this has a length of $(n - 1) + 1 + n$, that is $2n$, which is less than the heuristic value of the initial state so this action will be chosen as the one to apply.

The next state, however, is at the start of a plateau. The actions applicable (those for which all the preconditions are satisfied) are either to `drop` the ball that has been picked up, `pickup` another ball or `move` to the next room. The 'correct' action would be to `pickup` another ball: the relaxed plan to the goal state for the resulting state would be to `drop` one of the balls, `pickup` all the remaining balls in the newly-freed gripper, `move` to the next room, and `drop` all the balls. However, the heuristic value of this state would be $1 + (n - 2) + 1 + n$, or $2n$, the same value as the state in which the action is applied. Moving to the next room would produce a state with the heuristic value of $2n$ (`move` to the initial room, `pickup` remaining $(n - 1)$ balls, `drop` all balls in the final room—no `move` action is required to move back to any room the robot has already visited). Dropping one of the balls would also produce a state with a heuristic value of $2n$ (`pickup` all remaining $(n - 1)$ balls in newly-freed gripper, `move` to next room, `drop` all balls). As all successor states have the same RPG heuristic value as their parent state, the heuristic is unable to provide useful guidance as to which action to apply.

With some exhaustive search forward from this point, an improvement in heuristic value can be made in two ways: either `move` to the next room then `drop` a ball, or `pickup` a ball then `move` to the next room—both of these lead to heuristic values of $(2n - 1)$. The plateau will, however, be encountered each time the robot is in the first room, holding one ball, and the action choices are either to `pickup` another ball or `move` to the next room (or `drop` a ball). Each time the plateau is encountered, the action sequence to escape the plateau is identical—`move-drop` or `pickup-move` (in EHC the actual sequence chosen will depend on the order in which the actions are considered by the planner). Having to discover one of these action sequences by exhaustive search each time the plateau is encountered is a considerable bottleneck in the search process: this is true in general for many domains.

In order to address the overhead caused by recurrent plateaux in the search space, Marvin memoises the action sequences used to escape the previously encountered plateaux; these action sequences are used to form what are called 'Plateau-Escaping Macro-Actions'. A macro-action is generated from the action sequence using the code presented in Figure 4. Each step of the action sequence is considered in turn, and an abstract action step is made for it by replacing the entities given as parameters to the action with placeholder identifiers—one for each distinct entity. These placeholder identifiers then form the parameter list of the macro-action; and the recorded abstract action steps dictate the component actions from which the macro-action is built.

Returning to the `pickup-move` action sequence, the action sequence:

```
0: pickup robot1 ball2 room1
1: move robot1 room1 room2
```

would form a macro-action:

```
pickup-move (?a - robot) (?b - ball) (?c - room) (?d - room)
0:  pickup ?a ?b ?c
1:  move ?a ?c ?d
```

This macro-action can then be instantiated by specifying the parameters ?a to ?d, resulting in a sequence of actions. For example, (**pickup-move** `robot1 ball3 room1 room2`) would give an action sequence:

```
0: pickup robot1 ball3 room1
1: move robot1 room1 room2
```

In Marvin, the preconditions of the steps within the macro-action are not collected to give a single precondition formula for the macro-action. Instead, an instantiated macro-action is said to be applicable in a given state if the first component action of the macro-action is applicable, and subsequent actions are applicable in the relevant resulting states.

Having now built macro-actions from the plateau-escaping action sequences, when the search is later attempting to escape a plateau, these macro-actions are available for application. If the plateau arose due to the same weakness in the heuristic that led to an earlier plateau, then a macro-actions will be able to lead the search to a strictly better state by skipping over the intermediate states. The plateau-escaping macro-actions are only used when the search is attempting to escape a plateaux—this avoids slowing down search when the RPG heuristic is able to provide effective guidance using only single-step actions.

To reduce the number of macro-actions considered, and the blow-up in the size of the explored search space that would otherwise occur, the only macro-actions considered are those containing actions at the first time step that are helpful actions in the current state.

### 3.4 Macro-Actions in Use

The structure and reusability of macro-actions depends on the underlying topology of the problem space under the given heuristic function. When a problem space contains many similar occurrences of the same plateaux (which happens when a problem contains much repeating structure) the effort involved in learning macro-actions to escape these plateaux efficiently can be richly rewarded. In principle, the most benefit is obtained when the problem space features large, frequently recurring plateaux, since large plateaux are the most time-consuming to explore and the effort would need to be repeated on every similar occurrence. Short macro-actions (of two or three actions) indicate that the problem space contains small plateaux (although these might arise frequently enough for learned macro-actions to still be beneficial).

Problems with repeating structure include: transportation problems, where the same basic sequences of actions have to be repeated to move groups of objects from their sources to their destinations; construction problems, in which many similar components need to be built and then combined into a finished artifact; and configuration problems, in which multiple components of an architecture need to go through very similar processes to complete their functions, etc. The Dining Philosophers and Towers of Hanoi problems are good examples of problems with repeating structure.

Although using macro-actions during search has advantages—they can offer search guidance and allow many actions to be planned in one step—considering them during the expansion of each

```
 1: Procedure: BuildMacro
 2: parameters = [];
 3: parameter_types = [];
 4: abstract_steps = [];
 5: parameter_count = 0;
 6:
 7: for all action ?a in the action sequence used to escape a plateau do
 8:     abstract_parameters = [];
 9:
10:     for all parameter ?p of ?a do
11:
12:         if ?p ∈ parameters then
13:             index = parameter index of ?p in parameters;
14:             append (index) to abstract_parameters;
15:
16:         else
17:             parameters[parameter_count] = ?p;
18:             parameter_types[parameter_count] = type of ?p;
19:             append (parameter_count) to abstract_parameters;
20:             increment parameter_count;
21:         end if
22:     end for
23:     append (action type of ?a, abstract_parameters) to abstract_steps;
24: end for
25: return parameter_types and abstract_steps as a macro-action
```

Figure 4: Pseudo-code for building macro-actions from plan segments

state increases the branching factor. Thus, if a large number of unproductive macro-actions are generated the search space will become larger, making the problem harder, not easier, to solve. Whilst many of the plateau-escaping sequences are helpful in planning, some are specific to the situation in which they were derived, a situation which might not occur again in the plan. As macro-actions are learnt during the planning process—and there is no human intuition, or large test suite, to allow reusable macro-actions to be identified—care must be taken when deciding the points at which to consider their use in the planning process.

Plateau-escaping macro-actions are generated from situations in which the heuristic has broken down; therefore, the heuristic can be used as an indicator of when they are likely to be useful again during planning. As areas of repeating structure within the solution plan involve the application of similar (or identical) sequences of actions, they are likely to have similar heuristic profiles. In the case of plateau-escaping action sequences, the heuristic profile of the search landscape at their application is an initial increase (or no-change) of heuristic value, eventually followed by a fall to below the initial level—the profile occurring at a local minimum. If the plateau-escaping macro-actions are to be reusable, it is likely that the re-use will occur when the planning process is in a similar situation. As such, they are only considered for application in the exhaustive search step used to escape plateaux (both at the start or at any other point on a plateau).

Situations may arise where the use of macro-actions increases the makespan of the resulting plan due to redundant action sequences. For example, if in a simple game domain—with actions to move up, down, left or right— a macro-action is formed for 'left, left, left, left' and the optimal action sequence to escape a given plateau is 'left, left, left' then '{left, left, left, left}, right' may be chosen if the state reached by moving left four times is heuristically better than the one reached by applying a single-step 'left' action. Thus, macro-actions can have an adverse effect on plan quality.

Within the problem domains presented in IPC 4 (Hoffmann & Edelkamp, 2005) was the encoding of the Dining Philosophers problem, translated from Promela into a PDDL encoding. When solving this problem, two important macro-actions are formed: an eleven-step macro-action upon completion of the first period of exhaustive search; and a three-step macro-action upon completion of the second. The solution plan requires these macro-actions to be repeated many times, something which now—as a result of the macro-actions—involves simply applying a single action that results in a strictly better state. Without the macro-actions, the planning process consists of repeated episodes of exhaustive search to find the same two sequences of actions each time.

This behaviour can be seen in Figure 5 depicting the heuristic values of states generated with and without macro-actions, across the solution plan for the IPC 4 Dining Philosophers problem involving 14 philosophers. Initially, no macro-actions have been learnt so the search done by both approaches is identical. For the first 14 action choices the value of the heuristic, shown by the line in the graph, moves monotonically downwards as the planner is able to find actions to apply that lead to strictly better states.

After time step 14, the heuristic value begins to oscillate, at this point the planner has reached a plateau: there is no state with a strictly better heuristic value that can be reached by the application of just one action. As this is the first plateau reached, no macro-actions have been generated so the heuristic profiles are identical for both configurations. At time step 25 a state is reached that has a better heuristic value than that at time step 14. It is at this time that the plateau-escaping macro-action will be generated, memoising a lifted version of the sequence of actions that was used to escape the plateau. A brief period of search in which a strictly better state can be found at each choice point follows before the planner again hits a plateau.

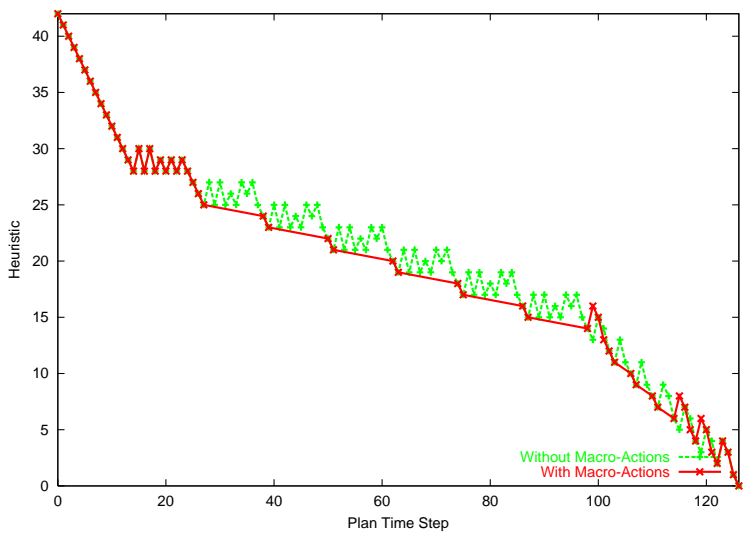

Figure 5: Heuristic landscape over makespan, with and without macro-actions.

The subsequent six plateaux consist of applying the same sequence of actions to six further pairs of philosophers; it can be seen that the heuristic fingerprints of the plateaux are identical. The version of Marvin in which macro-actions have been disabled repeats the expensive exhaustive search at each plateau: the heuristic value again goes through the process of increasing and then decreasing again before reaching a strictly-better state. The version using the plateau-escaping macro-actions, however, now has a single action to apply that achieves a strictly better state and search continues, stepping over the subsequent plateaux through the selection of macro-actions that yield strictly-better states.

When all of the larger plateaux have been overcome, a series of smaller plateaux are encountered. Again, it can be seen that for the first of these, both versions must complete a stage of exhaustive search; however, after the first of the smaller plateaux has been completed, the macro-action formed allows the subsequent plateaux to be bypassed. Finally, the plan finishes with a short previously unseen sequence of actions, where both versions must do exhaustive search.

## 4. Handling ADL

PDDL (McDermott, 2000) (the Planning Domain Definition Language) was first defined for use in the First International Planning Competition (IPC 1) at AIPS-98. Over the subsequent competitions, modifications have been made to the language as planning technology has evolved.

In the first three competitions, domains were available in which only STRIPS (Fikes & Nilsson, 1971) actions were used. STRIPS actions have conjunctive predicate preconditions, add effects, and delete effects defined in terms of action schema parameters and constant entities. To determine the preconditions and effects of a given ground action instance (an action whose parameters have been bound to specific entities) the action's parameters are substituted into the schema. For the action to be applicable in a given state, all of the precondition predicates must hold in that state; if the action is applied, a new state is generated from the previous state by removing all the predicates present in the delete effect list and adding those in the add effect list..

ADL action schemata (Pednault, 1989) extend the syntax of STRIPS action schemata. In ADL domains the language used to describe the preconditions of an action is extended to allow disjunctive, quantified and negative preconditions as well as the conjunctive preconditions that can be used in STRIPS domains. The syntax for describing the effects of actions is also extended to allow conditional effects—effects which are applied whenever a given condition holds.

The extended syntax provided by ADL not only increases the convenience with which a domain can be encoded, but can also reduce the size of the domain descriptions needed. For example, if an action schema has, as a precondition (or A B C) then, without ADL, three copies of the action schema would need to be made: one with a precondition (A), one with a precondition (B) and one with a precondition (C). If one is willing to tolerate such increases in domain-description size, and the number of objects in the domain is finite, it is possible to compile a given ADL domain and problem-instance pair into a domain-problem pair containing only STRIPS actions: in general, this compilation must be done once for each problem instance, not just once for each ADL domain. The ability to compile ADL domains into STRIPS domains was first demonstrated by the compilation procedure devised by Gazen and Knoblock (1997). Using these techniques in a preprocessing stage, FF is able to handle domains containing ADL actions whilst only reasoning about STRIPS actions internally. The output from FF's preprocessor stage was made available in IPC 4 to allow planners which could not handle ADL directly to solve compiled STRIPS formulations of the problems by loading a compiled domain-problem pair for each of the original problem instances in a given domain.

Whereas in previous competitions the ADL domains were simplified or manually reformulated to produce STRIPS domains, the STRIPS domains in IPC 4 were compiled automatically from ADL. The compilation used, based on the preprocessor of FF, results in a compiled domain-problem pair for each original problem instance. This compilation explicitly grounds many of the original actions, producing one compiled action schema (with preconditions and effects whose parameters refer to PDDL constants) per ground action that could arise in the original ADL problem. Whilst these compilations produce STRIPS domains that allow planning to be performed, they replace general action schemata with sets of specific action instances.

To allow the new features in Marvin to be used in the competition, Marvin was extended to include native support for ADL. By reasoning with the original ADL domain it is able to effectively abstract macro-actions from action sequences.

## 4.1 The Preconditions of ADL Actions

The preconditions of STRIPS actions consist of one structural feature - an 'and' clause, allowing conjunctive preconditions and predicates with constant or parameterised bindings. ADL actions have a far greater range of structural features in their preconditions. They allow 'or', 'imply', 'not', 'forall' and 'exists', which can be combined in any well-formed manner. In Marvin, the ADL preconditions are processed using two steps. First, all quantified preconditions are fully enumerated. Second, the resulting precondition formula is translated into Negation Normal Form (NNF) using the standard procedure: by replacing $(a \Rightarrow b)$ with $(\neg a \lor b)$, and using De Morgan's laws to eliminate negations of clauses. Further reductions are then applied to eliminate redundancy, such as replacing (and A (and B C)) with (and A B C), and (or A (or B C)) with (or A B C).

Internally, within Marvin, the NNF precondition formula forms the basis of a 'satisfaction tree', the nodes of which are the 'and' and 'or' elements of formula and the literals (negated and non-negated) form the leaves. The structure of the satisfaction tree for a given action schema is fixed, although the propositions at the leaves vary between groundings.

To determine which ground ADL action instances are applicable in a given state based on their preconditions, the algorithm shown by the pseudo-code fragment in Figure 6 is used. Initially, the satisfaction counters associated with each ground action's satisfaction tree nodes are reset using the following rules:

- Each 'and' nodes has its counter set to denote the number of children it has.

- Each 'or' node has its counter set to 1.

- Negative preconditions are assumed to be true, and the satisfaction counters of their parents decremented accordingly.

As these values are state-independent, for reasons of efficiency the values used to reset the satisfaction counters are computed once and cached for later use.

Having reset the satisfaction counters, each proposition in the current state is considered, and the satisfaction trees updated accordingly:

- The satisfaction counters of parent nodes that have the current proposition as a negative precondition are incremented.

- The satisfaction counters of parent nodes that have the current proposition as a positive precondition are decremented.

Then, by propagating the effects of truth value changes upwards through the tree, any action whose root node has sufficiently many children satisfied is applicable.

## 4.2 The Effects of ADL Actions

ADL extends the action definitions of STRIPS actions by allowing quantified and conditional effects. As in preconditions, the former are dealt with by enumeration; the latter are dealt with depending on their conditions.

If a conditional effect is dependent *only* on static predicates it is possible to determine when grounding an action whether or not it applies for that instance: the static information does not change from state to state. If the effect depends on dynamic predicates, it is necessary to consider, in each state, whether the effect applies. To achieve this, the effect and its conditions are used to form a sub-action. The sub-action has the conditional effect's condition as its preconditions, and the conditional effect itself as its effects. As conditional effects can be nested in the original operator schemata, the sub-actions themselves may also have conditional effects; in which case the sub-action-creation step is applied recursively, creating nested sub-actions as necessary.

The applicability of ground sub-actions in a given state is performed in the same manner as normal actions. When an action is applied, any sub-actions that are also applicable are applied alongside it, thereby preserving the conditional effects of the original operator.

## 4.3 Modifying the Relaxed Planning Graph

It is necessary to modify the Relaxed Planning Graph expansion and plan-extraction phases to make it possible to apply the heuristic when the domain contains ADL actions. Work has been done on

```
 1: Procedure: test-action-applicability
 2: reset_satisfaction_counters();
 3:
 4: for all predicate ?p in the state do
 5:
 6:     for all (ground action ?a, tree node ?c) pair having ?p as a negative precondition child node do
 7:         tree_node_to_update = ?c;
 8:
 9:         while tree_node_to_update is still valid do
10:             old_value = value in tree_node_to_update;
11:             value in tree_node_to_update = old_value + 1;
12:
13:             if value in tree_node_to_update > 0 && old_value = 0 then
14:                 tree_node_to_update = parent of tree_node_to_update;
15:
16:             else
17:                 tree_node_to_update = invalid;
18:             end if
19:         end while
20:     end for
21:
22:     for all (ground action ?a, tree node ?c) pair having ?p as a positive precondition child node do
23:         tree_node_to_update = ?c;
24:
25:         while tree_node_to_update is still valid do
26:             old_value = value in tree_node_to_update;
27:             value in tree_node_to_update = old_value -1;
28:
29:             if value in tree_node_to_update = 0 && old_value > 0 then
30:                 tree_node_to_update = parent of tree_node_to_update;
31:
32:             else
33:                 tree_node_to_update = invalid;
34:             end if
35:         end while
36:     end for
37: end for
38: applicable_actions = ∅;
39:
40: for all ground action ?a do
41:
42:     if root tree node is satisfied then
43:         add ?a to applicable_actions;
44:     end if
45: end for
```

Figure 6: Pseudo-code for action applicability testing

extending full graphplan planning graphs to reason with a subset of ADL actions (Koehler, Nebel, Hoffmann, & Dimopoulos, 1997); the approach taken in Marvin extends the relaxed planning graph structure to handle all of the available ADL constructs. The effect of the modifications is that the same heuristic estimate is obtained as if a precompiled STRIPS domain formulation was used.

When building a conventional relaxed planning graph the assumption is made that, in the first fact layer, all the facts present in the state to be evaluated are true and all other facts are, implicitly, false. Facts are then gradually accumulated by the application of actions, add effects adding facts to the spike (Long & Fox, 1999). Actions become applicable when their preconditions are all present; i.e. they have all been accumulated. The STRIPS actions used to build a conventional relaxed planning graph necessarily have no negative preconditions, so it is sufficient to consider when facts have a positive truth value and determine action applicability from this. ADL actions, however, can also have negative preconditions, corresponding to facts which must be false. Within a conventional relaxed planning graph, no record is made of whether it is possible for a given fact to have a negative truth value.

To handle negative facts within the relaxed planning graph used in Marvin, a second spike is added. As with the positive-fact spike, all the facts present in the state to be evaluated are true and all other facts are, implicitly, false. However, unlike the positive-fact spike, facts are then gradually *eroded* by the applications of actions; with their delete effects marking the fact in the negative-fact spike as having been deleted. The inherent relaxation on which the relaxed planning graph is founded is still preserved, though: delete effects have no effect on the positive-fact spike; and, similarly, add effects have no effect on the negative-fact spike.

If a precompiled STRIPS domain formulation was used, additional complimentary propositions are added to denote when each proposition is not true. These accumulate alongside the original domain propositions, and in this way are able to satisfy negative preconditions. The negative fact spike, as discussed, has the same effect, although rather than recording which propositions are available in a negated form at each layer, it records which propositions are not available in a negated form.

As discussed, ADL action preconditions are preprocessed such that negation is only applied to the leaves of the satisfaction tree; i.e. only applied to unit facts forming part of the actions' precondition structures. Within the relaxed planning graph a given fact leaf can now be marked as satisfied if either one of the following holds:

- It is a positive fact leaf, and the fact contained therein has been added to the positive-fact spike.

- It is a negative fact leaf, and the fact contained therein has either never been in the negative-fact spike or has since been marked as deleted.

Plan graph construction proceeds in a manner similar to that used to build a conventional relaxed planning graph. Each of the newly present or newly deleted facts are considered in turn, and their effects on the applicability of all available actions noted. Should the updating of the satisfaction tree of an action lead to it becoming applicable:

- The action is added to the action spike, available at the next fact layer.

- Previously unseen add effects are added to the positive-fact spike, available at the next fact layer.

- Delete effects deleting a fact still present in the negative-fact spike mark the fact as being deleted and available to satisfy negative preconditions from the next fact layer.

For efficiency, the first action to achieve each fact is stored when it is added to the positive-fact spike, along with the earliest layer at which that action is applicable. Similarly, the first action that deletes each fact that has ever been in the negative-fact spike is noted. Relaxed plan extraction consists of regressing through the layers of the relaxed planning graph, selecting actions that achieve the goals that are to be achieved at each layer. Initially, each proposition in the goal state is added to the layer of goals for the layer in which it first appears (or disappears, in the case of negative goals). To extract a plan, the next goal is repeatedly taken from the deepest action layer with outstanding goals. Its first achieving action is added to the plan and its preconditions, taken from its satisfaction tree, are added to the goals for the first layer in which they appear. The process finishes when there are no more outstanding goals at any layer. If a sub-action (that is, an action created to represent the conditional effect of an ADL action, see Section 4.2) is chosen to achieve a given proposition, the preconditions of its parent action(s) are also added to the goals for the first layer in which they appear.

When considering adding the preconditions of an achieving action to the layer in which they appear, a collection of disjunctive preconditions may arise. In this situation, the first satisfied precondition or negative precondition in the disjunction is added as a subgoal in an earlier layer. This avoids adding many redundant actions to satisfy each of a the disjunctive preconditions, where only one needs to be satisfied. The precondition chosen to be satisfied from each collection of disjunctive preconditions is the first for which an achiever was found when building the relaxed planning graph, thus providing the same heuristic estimate as if the compiled STRIPS domain formulation was used. In the compiled STRIPS domain formulation, the disjunctive precondition would give rise to several action instantiations; the first applicable of these would be chosen as the achiever for the desired fact.

At the start of the planning process, a relaxed planning graph is constructed forward from the initial state. However, rather than stopping when the goal literals appear, graph construction stops when no more ground actions become applicable. The actions and propositions appearing in this relaxed planning graph are a superset of all the actions and propositions appearing in later relaxed planning graphs: these actions and propositions discovered are then used to form a cache detailing the proposition–action dependencies. Using this cached information, the code shown in Figure 6 can be used to determine the actions applicable in a given state, and the relaxed planning graphs used to calculate heuristic values can be extracted more efficiently.

## 5. Handling Derived Predicates

In IPC 4, PDDL was further extended with the addition of *Derived Predicates* (Hoffmann & Edelkamp, 2005). Derived Predicates, used in three of the competition domains, allow higher-level concepts to be recursively derived from other propositions. These derived predicates can then be present in the preconditions of actions, and allow higher-level concepts in the domain to be reasoned with. For example, in the BlocksWorld domain, the derivation rule for the 'above' predicate is as follows:

```
(:derived (above ?x ?y)
    (or (on ?x ?y) (exists (?z) (and (on ?x ?z) (above ?z ?y)))))
```

Should a planner not include native support for derived predicates, it is possible to compile domains containing derived predicates into "flattened" domains that do not. However, it is not possible to do this without a super-polynomial increase in the size of the domain and the solution plan (Nebel, Hoffmann, & Thiebaux, 2003). At IPC 4, compiled versions of the domains that contained derived predicates were made available for competitors who could not support derived predicates. However, the sizes of the problems that could be compiled were restricted by the concomitant sizes of the PDDL files produced by the compilation process and the computational effort necessary to solve the compiled problems.

IPC 4 was the first planning competition to make use of derived predicates in its domains. As it has been shown that derived predicates cannot be reasoned about efficiently through compilation (Nebel et al., 2003) steps were taken to provide native support for them in Marvin.

It is also possible to compile derived predicates appearing in domains by adding actions to instantiate the derived predicates on an as-needed basis (Gazen & Knoblock, 1997). Using this compilation, the 'above' derivation rule from the blocksworld problem described above would be compiled to the following action:

**confirm_above ?x ?y**
    pre: (or (on ?x ?y) (exists (?z) (and (on ?x ?z) (above ?y ?z))))
    add: (above ?x ?y)

If this is to be used as a domain compilation, each of the original actions in the domain must be extended to delete all of the 'above' propositions, forcing the **confirm_above** actions to be used to re-achieve the 'above' preconditions for any action that requires them. In this case, each action is given the additional effect:

    (forall (?x ?y) (not (above ?x ?y)))

Although effective in STRIPS domains, it is not possible to use such a compilation for domains making use of negative preconditions as the re-derivation of derived predicates occurring as negative preconditions of actions is not enforced. For example, an action could be applied that modifies the 'on' propositions, leading to a state from which a number of additional 'above' properties could be derived. Deleting the 'above' propositions is a necessary step, as the confirm actions should re-assert any derived predicate for any action that needs it. However, when (above ?x ?y) is deleted, (not (above ?x ?y)) is true, and can be used as an action precondition. To deal with this issue it is necessary to prevent any non-confirm actions from being applied until all possible derived predicates have been re-asserted; this prevents actions from being applied when a given 'not above' is only temporarily true, i.e. whilst it has not yet been re-derived. To force the re-derivation of derived predicates, further dummy predicates and actions must be added to the domain. The necessary compilation results in a large increase in the size of the search space explored, and the additional dummy actions affect the usefulness of the relaxed-planning-graph heuristic.

The problems with using the Gazen & Knoblock compilation arise solely because, in its original form, it does not force all applicable confirm actions to be applied after each original action is applied. As such, if a planner generates the confirm actions internally and then deals with them appropriately, the compilation can still form the basis of an effective means for handling derived predicates.

To this end, when presented with a domain containing derived predicates, Marvin machine-generates the confirm actions and extends each (original) action to delete the derived predicates, as described. After each action is applied, all propositions that can be directly or recursively derived from the resulting state are instantiated by applying all applicable confirm actions. Along with avoiding an unwieldy compilation in domains with negative preconditions, handling the confirm actions internally in this manner provides performance improvements for two further reasons:

- As the confirm actions are automatically applied when appropriate, Marvin does not have to do search and perform heuristic evaluation to discover that the next action required will be a confirm action.

- Confirm actions are included alongside normal actions in the relaxed planning graph built for each state, but if used in the relaxed plan they do not contribute towards the heuristic value taken from its length, eliminating any noise they would otherwise add.

## 6. Results

The planning competition offers a great opportunity for assessing the relative performance of various techniques used in planning over a wide range of problems. Inevitably there will, however, be features that are not tested by the set of domains used in the competition. There will also be some domains in which many of the features of a planner collaborate to produce good results, rather than the results being directly attributable to one individual feature. Here we discuss the results from the competition and present further results to clarify which of the features of Marvin contribute to the performance in each particular case.

It is important to note that when we refer to macro-actions generated and used by Marvin these are all generated during the planning process for that specific problem. No additional learning time or knowledge gained from solving other problems was used by Marvin in the competition, or in producing the additional results presented in this paper. Although some planners can use additional 'learning time' when solving a series of problems, a satisfactory way to incorporate this extra time into the time taken to solve each problem, as measured in the planning competition, has yet to be found. In the planning competition the planners are compared based on their performance on isolated problem instances, which is still an interesting comparison to make.

The results presented were produced on two machines: a machine at the University of Strathclyde (with a 3.4GHz Pentium 4 processor) and the IPC 4 competition machine (with a 3GHz Xeon processor). In both cases, the planner was subjected to a 30 minute time limit and a 1Gb memory usage limit. All results that are directly compared against each other (i.e. appear on the same graph) are produced on the same machine. The domains used for evaluation are taken from IPC 3 and IPC 4, and are described in detail in the papers giving an overview of each of the two competitions (Long & Fox, 2003; Hoffmann & Edelkamp, 2005).

### 6.1 Plateau-Escaping Macro-Actions

To assess the effect of plateau-escaping macro-actions on planner performance, tests were run across a range of planning domains with macro-actions enabled and disabled. The results of these tests are shown in Figures 7 and 8, illustrating the time taken to find a solution plan and the makespan of the plan found respectively.

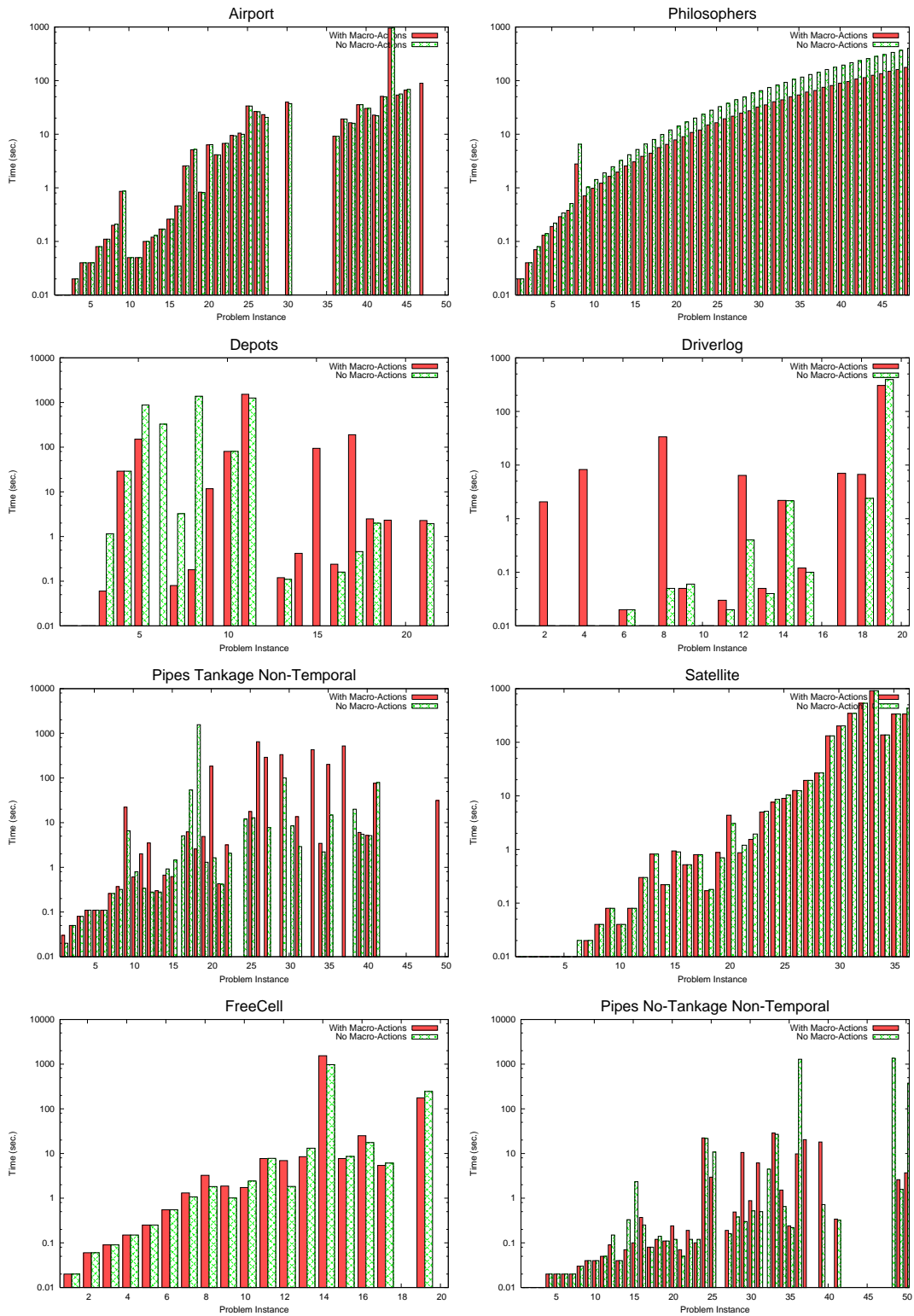

Figure 7: CPU time showing the results of planning with and without plateau-escaping macro-actions on a range of domains (from left to right: Airport, Philosophers, Depots, Driverlog, Pipestankage-nontemporal, Satellite, FreeCell, Pipesnotankage-nontemporal).

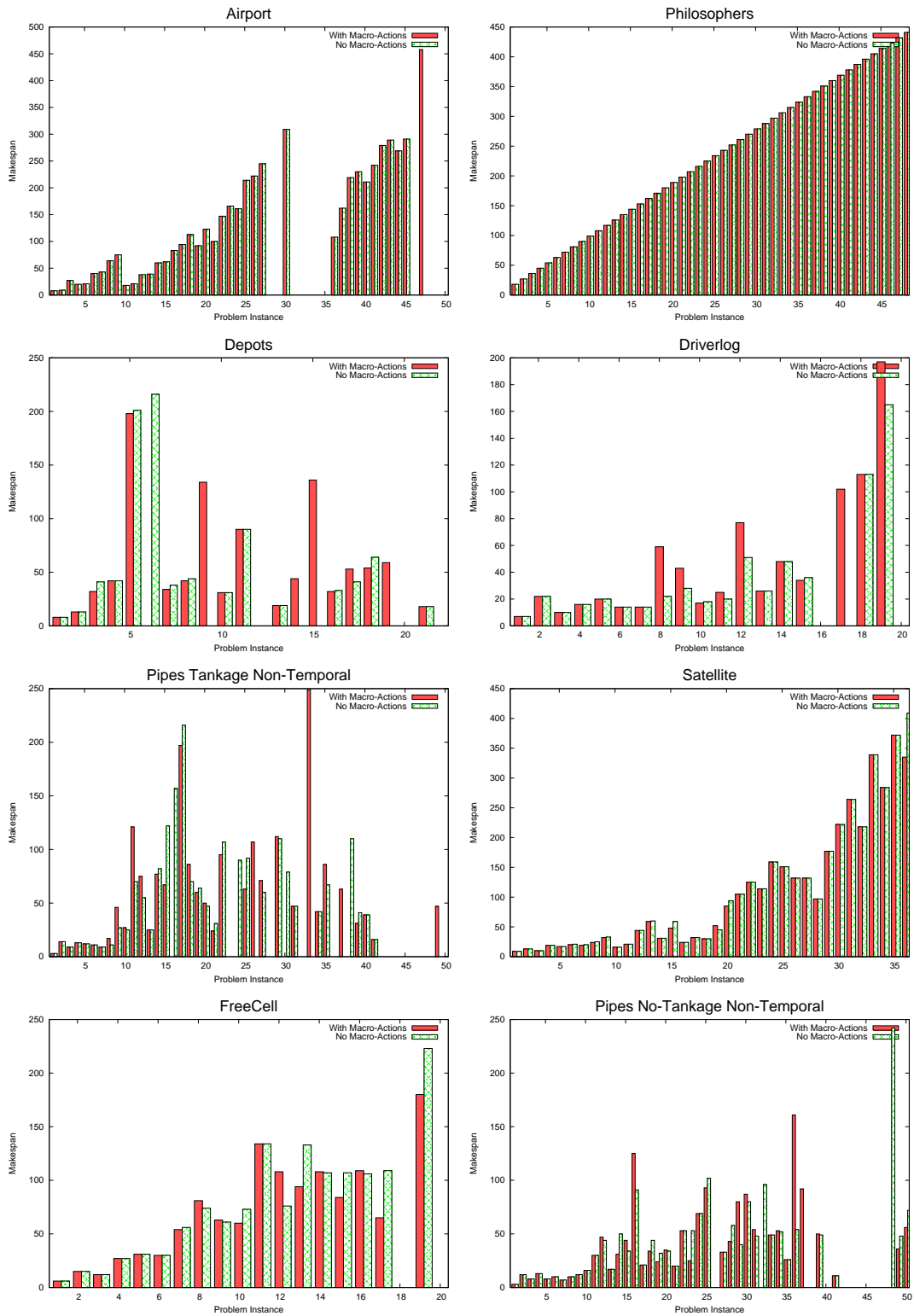

Figure 8: Makespan of the solution plans found when planning with and without plateau-escaping macro-actions on a range of domains (from left to right: Airport, Philosophers, Depots, Driverlog, Pipestankage-nontemporal, Satellite, FreeCell, Pipesnotankage-nontemporal).

In the Airport domain the time taken to find plans and the makespans of the plans found were almost identical. A strictly better successor can usually be found to each state when using EHC, and it is clear in this domain that the addition of macro-actions from the occasional plateau has not degraded the performance of the planner. The performance of the two configurations deviates at problem 47, where planning with macro-actions was able to find a solution plan but planning without macro-actions was not. Closer inspection of the output from the planner reveals that in this case, some way into EHC search, a plateau is encountered and escaped; in the configuration using macro-actions, this leads to the formation of a macro-action. Later in search, another plateau is encountered. At this point, the earlier macro-action can be used to lead to a strictly better state, from which a solution plan can ultimately be found using EHC. If the macro-action is not available, however, the sequence of actions found to escape the plateau leads to a different exit point, from which a solution plan cannot be found using EHC.

In the Philosophers domain neither the makespans of the plans found nor the coverage differs between the two configurations tested. Using macro-actions, however, leads consistently to improved performance as the plateaux encountered during search require the application of the same action sequence. Consistently, across the problems, searching with macro-actions is faster by a factor of two; and furthermore, the factor is increasing with problem size, suggesting it has better scalability.

In the Depots domain, using macro-actions improves coverage, allowing 18 problems to be solved within the time limit rather than 15. Further, in many cases, the time taken to find a plan is reduced. In one case, problem file 6, planning without macro-actions is able to find a plan where planning with macro-actions cannot. Here, planning without macro-actions is unable to find an exit point from one of the plateaux encountered later in search, and resorts to best-first search. Planning with macro-actions, however, is able to reach a greater number of successor states from the nodes on the plateau and is unable to exhaust the reachable possibilities and terminate EHC search within the 30-minute time limit.

In the Driverlog domain, using macro-actions generally increases the time taken to find plans and has an adverse effect on the makespan. In this domain, macro-actions containing varying-length action sequences consisting of repeated `walk` or `drive` actions are inferred. In practice, these are detrimental in two ways: they have a large number of instantiations and dramatically increase the branching factor, reducing performance; and they are only usefully reusable in situations where the prescribed number of `walk` or `drive` actions are needed. Despite this, planning with macro-actions is able to find solution plans in 18 of the problems, whereas planning without the macro-actions is only able to solve 17 of the problems. In the problem in question, problem 17, the increased number of successor states visible from the nodes on plateaux due to the presence of macro-actions allows EHC to find a solution plan rather than resorting to best-first search, which would ultimately fail within the time limit set.

In the Pipestankage-nontemporal domain, it is not clear at first whether macro-actions are beneficial or not. The number of problems solved by both configurations is the same, 34, and the impact on makespan appears to be insignificant, improving it in some cases but making it worse in others. However, looking at the harder problems from problem 25 upwards, planning with macro-actions is able to solve 13 rather than 11 problems, suggesting it is able to scale better to larger problems compared to searching without macro-actions.

In the Satellite domain both configurations exhibit similar performance, in terms of both the time taken to find a solution plan and the makespan of the plan found, as the relaxed planning graph heuristic is generally able to provide good search guidance. The exception is problem 36: here, the inference of a macro-action allows search to be completed using EHC rather than resorting to best-first search, reducing the time taken to find a plan.

In the FreeCell domain, macro-actions appear to lead to improved makespans and have negligible impact on the time taken to find solution plans. Intuitively, however, in a strongly directed search space (such as that in FreeCell, where it is possible to move a card from one location to another but often not to move it back) using a non-backtracking search strategy such as EHC should reduce the effectiveness of macro-actions, as the introduction of redundant action steps as part of a macro-action instantiations can lead search towards unpredicted dead-ends. The illustrated results, contradicting this intuition, can be ascribed to the nature of the FreeCell problems used in IPC 3. The problem files all have the four suits of cards, and from problem file 7 upwards have four free cells. The number of cards in each suit and the number of columns are gradually increased from 2 to 13 and 4 to 8 respectively. The effect of this, however, is that all but the hardest problems have a favourable free cells to cards ratio. When macro-actions are used, the impact of needlessly moving a card into a free cell is not significant as there is a generous allocation of free cells compared to the number of cards that might need to be stored there.

To provide a more reasonable test of whether macro-actions are beneficial in the FreeCell domain, twenty full-sized problem instances were generated and tests run to compare the performance of Marvin with and without macro-actions on these problems. The results of these tests can be seen in Figure 9 - clearly, the number of problems solvable within the 30 minute time limit and, generally, the time taken to find a solution plan is improved when macro-actions are not used.

In the Pipesnotankage-nontemporal domain the results obtained do not show a significant advantage or disadvantage to using macro-actions: the planner is faster for some of the problems when using macro-actions, but is slower on others; similarly, the planner produces plans with shorter makespans on some problems when using macro-actions, but longer makespans on others. Two results are obtained when macro-actions are not used that are very close to the 30-minute cut-off. The first of these is solved in around 10 seconds when macro-actions are used; the second can be solved using macro-actions if an extra 5 minutes of CPU time are allowed, or if a slightly faster computer is used.

Overall, it can be seen that the effect of plateau-escaping macro-actions on the execution time of the planner varies depending on the domain in question:

- In the Philosophers, Depots, Driverlog and Pipestankage-nontemporal domains, the use of macro-actions improves the performance of the planner, either in terms of coverage or a general reduction in the time taken to find solution plans.

- In the FreeCell domain, worse performance is observed when macro-actions are used.

- In the Airport, Pipesnotankage-nontemporal and Satellite domains the difference in performance is minimal.

Furthermore, with the exception of the Driverlog and FreeCell domains (where the makespan of solution plans is generally increased when using macro-actions) the use of macro-actions does not significantly affect the makespan.

## 6.2 Greedy Best-First Search versus Best-First Search

To assess how the performance of greedy best-first search compares to conventional best-first search, we ran tests across a range of planning domains with EHC and macro-actions disabled to isolate the effect of the greedy best-first search approach. Overall, when analysing the results, it was observed that the choice of best-first search algorithm had little impact on the performance of the planner.

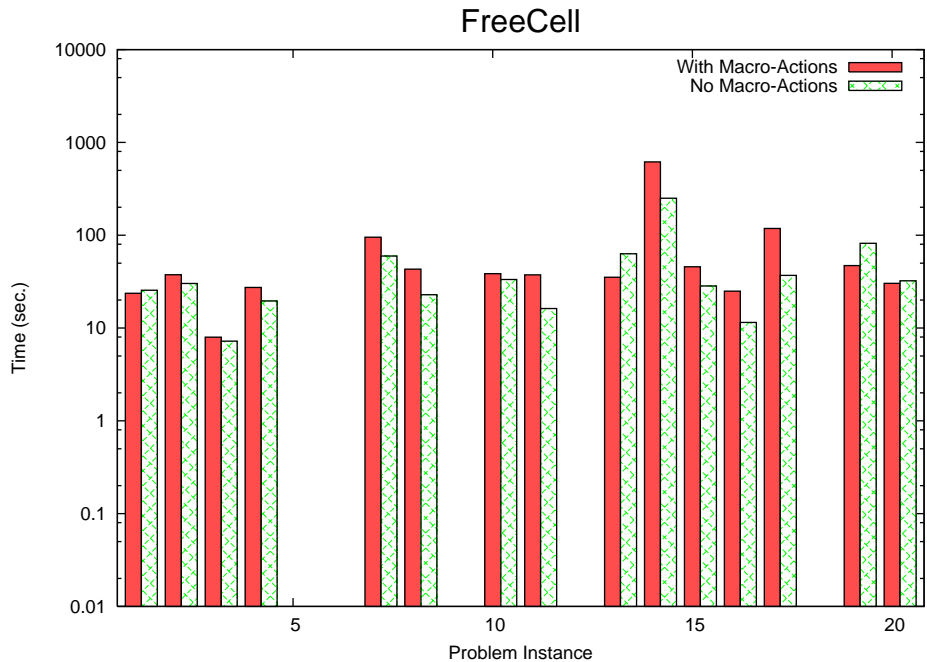

Figure 9: Time taken to solve twenty full-sized problems in the FreeCell domain, with and without plateau-escaping macro-actions.

### 6.3 Least-Bad-First Search versus Breadth-First Search

To assess the effect of using least-bad-first search rather than breadth-first search to escape plateaux in EHC search, we ran tests across a range of planning domains using each of the two search algorithms. The results of these tests are shown in Figures 10 and 11, illustrating the time taken to find a solution plan, and the makespan of the plan found.

In the Airport domain, plateaux arise in one of two cases:

- An unforeseen deadend has been reached; as no backtracking is made over action choices exhaustively searching the plateau is inexpensive, and EHC terminates rapidly.

- A short plateau has been reached, requiring two actions to be applied to reach a state with a strictly better heuristic value—here, the two actions found by both least-bad-first and breadth-first search were identical.

As can be seen from the planning time and makespan graphs, using least-bad-first search rather than breadth-first search has no impact on planning time or solution plan quality in the Airport domain: the time spent searching plateaux is negligible, and the escape paths found are identical under the two plateau-search approaches.

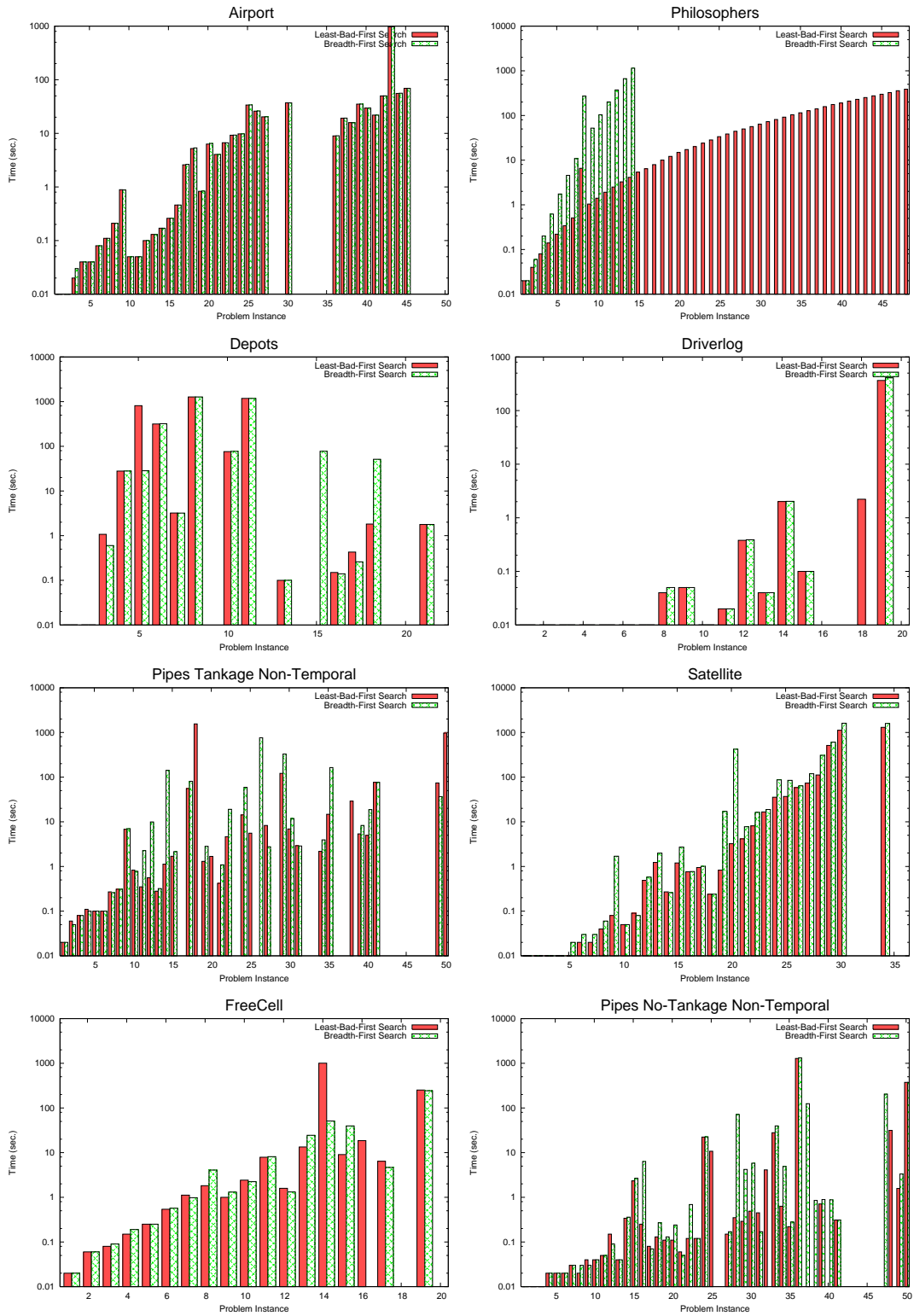

Figure 10: CPU time showing a comparison between using breadth-first and least-bad-first search on plateau search on a range of domains (from left to right: Airport, Philosophers, Depots, Driverlog, Pipestankage-nontemporal, Satellite, FreeCell, Pipesnotankage-nontemporal). These results were generated without using macro-actions.

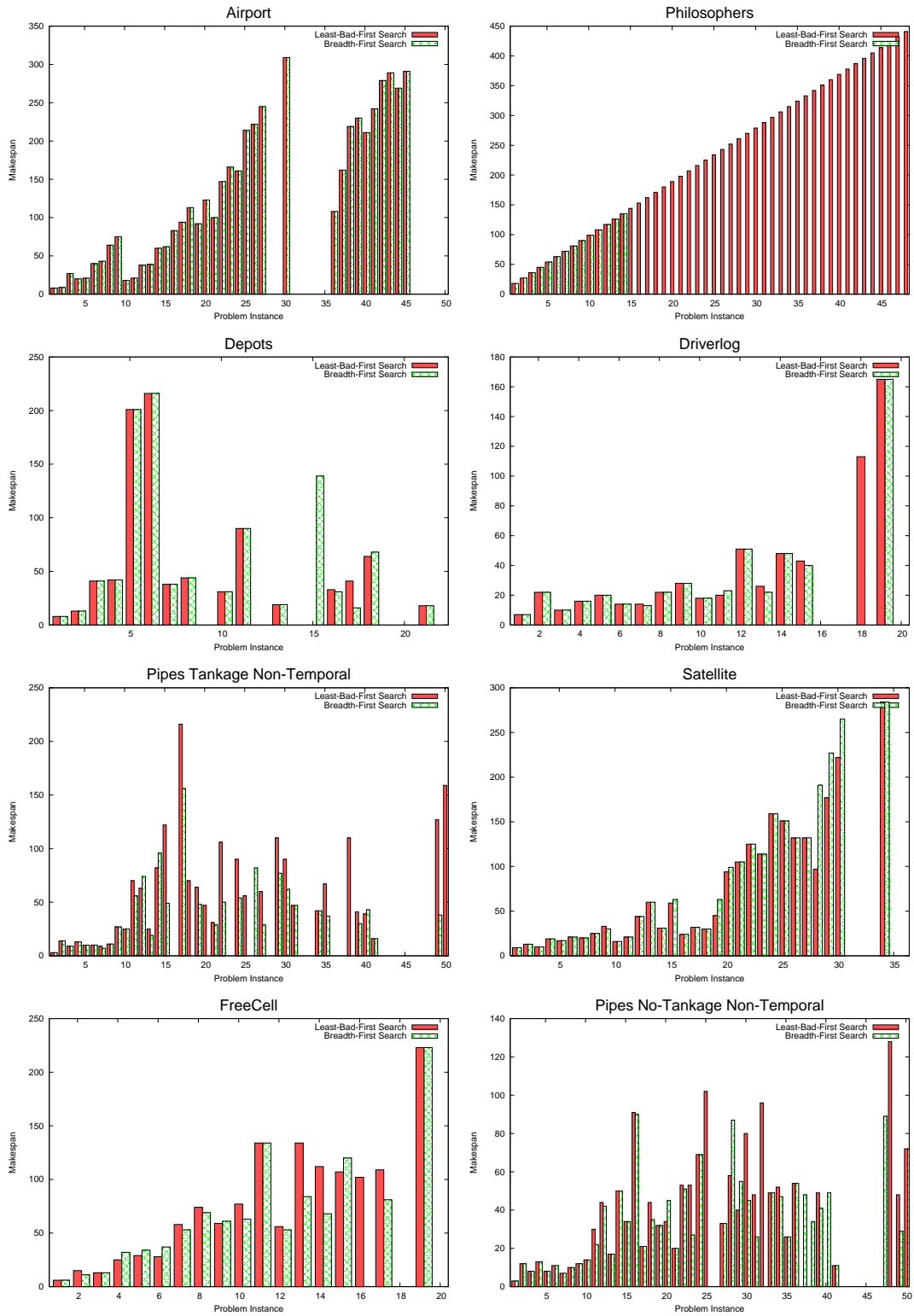

Figure 11: Makespan of plans produced using breadth-first and least-bad-first search during plateau search on a range of domains (from left to right: Airport, Philosophers, Depots, Driverlog, Pipestankage-nontemporal, Satellite, FreeCell, Pipesnotankage-nontemporal). These results were generated without using macro-actions.

In the Philosophers domain, search time is dramatically reduced by using least-bad-first search rather than breadth-first search on plateaux. Using least-bad-first search, all 48 problems are solved; using breadth-first search, only the first 14 are solved. The plans found in the first 14 have identical makespans, although the actions occur in differing orders in the two plans.

The search landscape provides some insights into why least-bad-first search is suited to this problem domain. At the start of the largest plateaux encountered, each action leads to a state with a strictly worse heuristic value; each of these corresponds to applying the action 'queue-write' to a philosopher. From each of these, a state with a less-bad heuristic is visible. When using least-bad-first search, this less-bad state is considered before the others in the queue, avoiding the redundant search that would otherwise be performed by breadth-first search. Adding more philosophers to the problem causes a dramatic increase in the amount of redundant search performed when breadth-first search is used, leading to the observed performance improvement when a least-bad-first approach is taken.

In the Depots domain, we can observe the effect of differing exit points to plateaux when using least-bad-first and breadth-first search. When solving problem 18, least-bad-first search is able to solve the problem in substantially less time: EHC search is able to escape all the plateau encountered, and find a solution plan. Breadth-first search, however, leads to the termination of EHC, and exhaustive best-first search being used. On problem 15, however, breadth-first search is able to find a solution plan where least-bad-first search cannot; also, problem 5 is solved in much less time. In these two cases, it is not the success of breadth-first search on plateaux which leads to the improved performance, but its failure; EHC search terminates and resorts to best-first search in less time when breadth-first search is used than when least-bad-first search is used.

In the Driverlog domain, one additional problem, number 18, can be solved when least-bad-first search is used instead of breadth-first search. EHC using breadth-first search leads to a plateau which cannot be escaped, and EHC aborts without a solution plan; the resulting exhaustive best-first search cannot be completed within the allowed 30 minutes. The makespans of the plans found by the two approaches do not differ significantly.

In the Pipestankage-nontemporal domain, it can be seen that the use of least-bad-first search generally reduces the time taken to find solution plans. 34 problems are solved when using least-bad-first search compared to 30 when using breadth-first search and, in the majority of cases, the time taken to find a solution plan is reduced. The makespans of the resulting solution plans are generally increased when least-bad-first search is used, though, as the suboptimal exit paths found in this domain are often longer than the (optimal-length) paths found when breadth-first search is used.

In the Satellite domain using least-bad-first search leads to a reduction in planning time and, in many cases, a reduction of the makespan. In particular, the performance on problems 19 and 20 is substantially improved. The makespans on problems from 28 to 30 inclusive are also improved.

On the twenty standard benchmark FreeCell problems using least-bad-first search allows one additional problem to be solved within the 30 minute time limit. As with the results obtained when assessing the impact of macro-actions on planner performance, we obtained a more interesting and useful set of data. Figure 12 shows the results of these experiments: it can be seen that although least-bad-first search often improves the time taken to solve problems, the coverage overall is reduced, and no additional problems are solved where they previously were not.

In the Pipesnotankage-nontemporal domain, one additional problem can be solved using breadth-first search rather than least-bad-first search. Also, in many cases, the use of least-bad-first search increases the makespan of the solution plan found. Overall, although time reductions can occur when solving some problems when using least-bad-first search, the use of breadth-first search provides better overall performance both in terms of planning time and makespan.

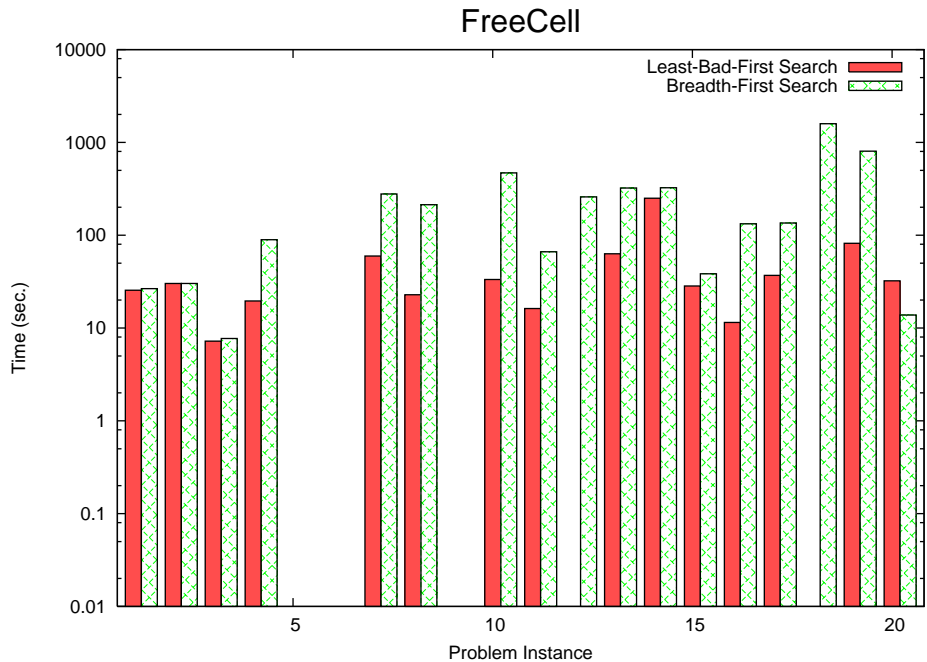

Figure 12: Time taken to solve twenty full-sized problems in the FreeCell domain, with least-bad-first and breadth-first search on plateaux (without macro-actions).

Overall, it can be seen across the evaluation domains that the performance of the planner when using least-bad- or breadth-first search varies, in terms of planner execution time and plan quality:

- In the Philosophers domain, the use of least-bad-first search provides a substantial improvement in planner performance.

- In the Satellite, Driverlog and Pipestankage-nontemporal domains, the execution time of the planner is generally improved by the use of least-bad-first search (with some reduction in plan quality in the latter of these).

- In the Airport and Depots domain, the impact on performance is minimal, either in terms of execution time or solution plan quality.

- In the FreeCell and Pipesnotankage-nontemporal domains, performance of the planner is degraded, both in terms of execution time and plan quality.

### 6.4 Handling Derived Predicates

It is possible to reason with domains involving derived predicates by precompiling the domain, adding additional actions to support the derived predicates, and then planning in the usual manner (see Section 5). The necessary compilation, however, causes a large increase in the size of the domain. If the planner performs the compilation itself, generating the confirm actions and segregating them from the normal actions internally, it can avoid the search overhead the compiled domain would incur.

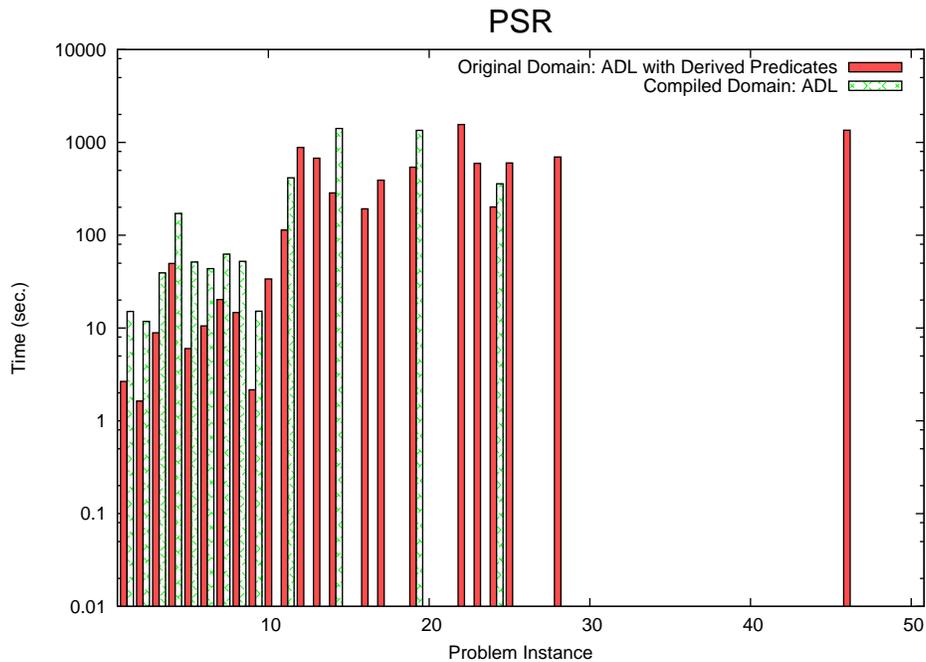

Figure 13: Time taken to solve problems in the PSR domain with and without Derived Predicates.

Three IPC 4 domains make use of derived predicates: PSR (Power Supply Restoration), Philosophers and Optical Telegraph. To assess the impact the native support of derived predicates was having on planner performance, tests were run in these domains using the original domains containing derived predicates, and using the compiled domains. The results of these tests are shown in Figures 13, 14 and 15.

In the PSR domain, the support of derived predicates substantially reduces the time taken to find solution plans. This improvement in efficiency allows 23 rather than 12 problems to be solved within the 30 minute time limit.

Marvin is only able to solve a few of the problems in the promela/optical-telegraph domain. On the smaller problems, the performance is better without derived predicates; nonetheless, two of the larger problems (problems 8 and 9) can be solved when working with the original domain where previously they could not, and overall one additional problem is solved with derived predicates.

In the Philosophers domain, supporting derived predicates natively yields substantial reductions in planning time. Using the compiled ADL domain formulation, only the first nine problems can be solved. With native derived predicate support, all 48 problems can be solved.

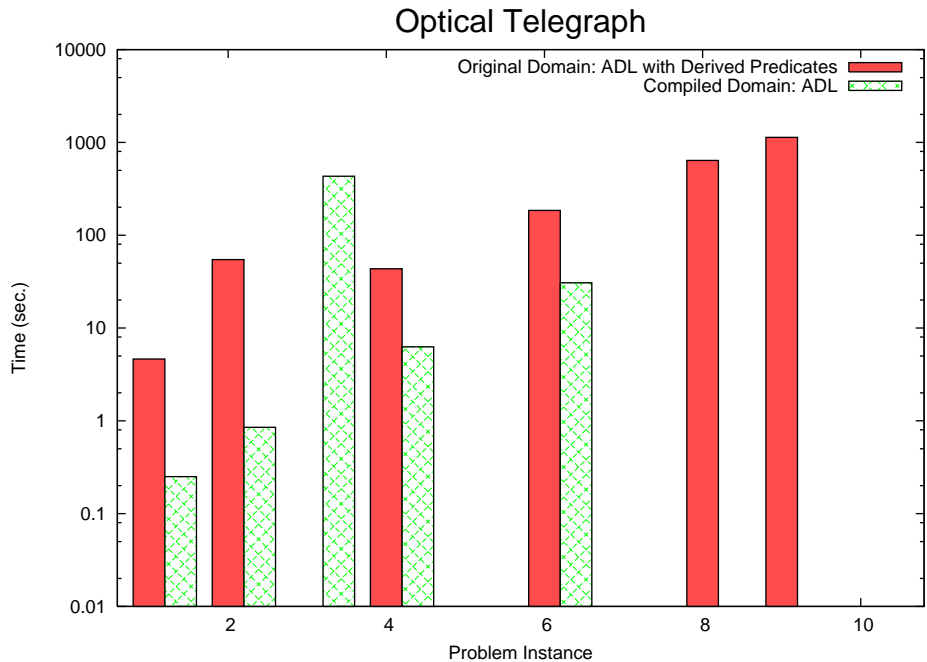

Figure 14: Time taken to solve problems in the Optical Telegraph domain with and without Derived Predicates.

## 6.5 Native ADL Support

The native support of ADL in Marvin provides two benefits, arising from the ability to use non-compiled domain formulations:

- Potentially improved efficiency, due to a more-efficient representation.

- The ability to infer reusable, parameterised macro-action sequences from the original ADL actions, whose parameters are lost as a side-effect of the process used to compile ADL to STRIPS domains.

### 6.5.1 The Effects of Using a Non-Compiled Domain

To assess the effect of native support for ADL constructs on the performance of Marvin, we ran a series of tests comparing the planner's performance when given both the STRIPS and ADL domain encodings. Macro-actions were disabled in both cases to isolate the effect the encoding itself was having on performance. In IPC 4, ADL was used to encode four of the domains: Airport, Philosophers, Optical Telegraph and PSR. STRIPS compilations were made available for each of these domains, in which each ground action that could arise when using the original ADL domain was made into a fixed-parameter STRIPS action. In the Philosophers, Optical Telegraph and PSR domains, the domain formulations making use of Derived Predicates were used.

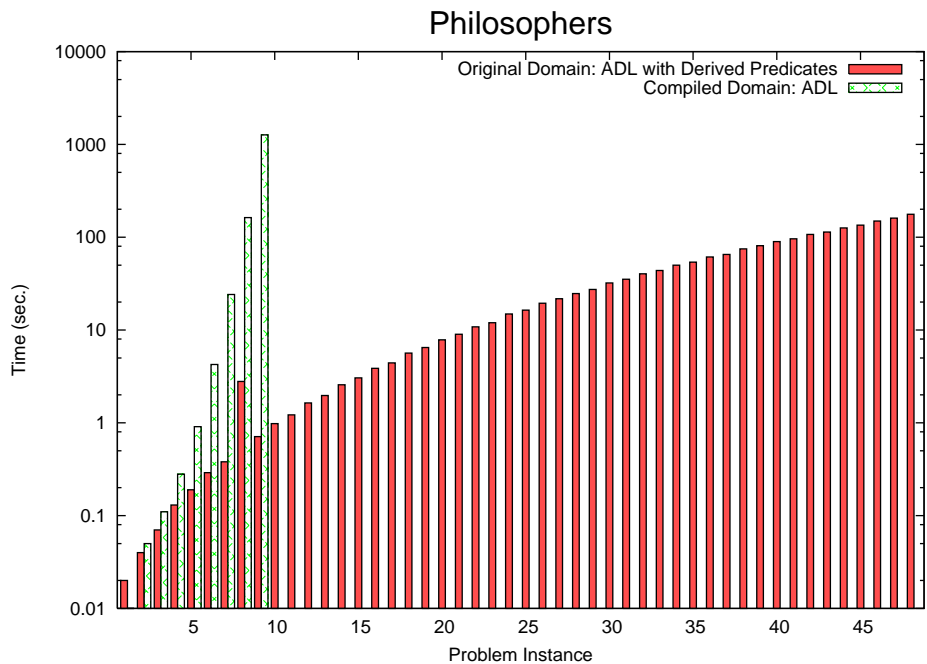

Figure 15: Time taken to solve problems in the Philosophers Domain with and without Derived Predicates.

In the Airport, Optical Telegraph and PSR domains, the performance of Marvin (with macro-actions disabled) was unaffected by the use of either the ADL or STRIPS domain encoding. The ADL domain encodings did not give rise to inefficient compiled STRIPS encodings.

In the Philosophers domain, the use of the ADL domain encoding resulted in a reduction in planning time when compared to the use of the compiled STRIPS encoding. As can be seen in Figure 17, more problems can be solved within the 30 minute time-limit if the ADL encoding rather than the STRIPS encoding is used, even disregarding the improvements in performance provided by the use of macro-actions.

### 6.5.2 THE EFFECTS OF INFERRING MACRO-ACTIONS

Supporting ADL natively in Marvin allows lifted macro-action schemata to be inferred during search: in the compiled STRIPS domain formulations presented in IPC 4, the actions in the plateau-escaping action sequences have few or no parameters, removing the opportunity to infer parameterised action sequences to use as the basis for macro-actions. Reusable macro-actions can be inferred in STRIPS domains, as in many of the domains discussed in Section 6.1; but the compilation from ADL to STRIPS produces a domain in which the macro-actions cannot, in practice, ever be reused.

To assess the effects of plateau-escaping macro-actions when using the ADL domain formulation, tests were run in the Philosophers, Optical Telegraph and PSR domains using the ADL domain formulation with macro-actions enabled and disabled. Results for the Airport domain are presented in Section 6.1, and results in the other domains will now be discussed.

0: (activate-trans philosopher-1 philosopher forks–pid-wfork state-1 state-6) [1]
1: (activate-trans philosopher-2 philosopher forks–pid-wfork state-1 state-6) [1]
2: (activate-trans philosopher-3 philosopher forks–pid-wfork state-1 state-6) [1]
3: (activate-trans philosopher-4 philosopher forks–pid-wfork state-1 state-6) [1]
4: (activate-trans philosopher-0 philosopher forks–pid-wfork state-1 state-6) [1]

**5: Macro-Action A Derived Here, using philosopher-4, philosopher-3, forks-4 and forks-3**

16: (activate-trans philosopher-3 philosopher forks–pid-rfork state-6 state-3) [1]

**17: Macro-Action A, using philosopher-2, philosopher-1, forks-2- and forks-1**

28: (activate-trans philosopher-1 philosopher forks–pid-rfork state-6 state-3) [1]

**29: Macro-Action B Derived Here, using philosopher-3 and -forks-3-**

32: (activate-trans philosopher-3 philosopher forks-_-pidp1_11_-rfork state-3 state-4) [1]

**33: Macro-Action B, using philosopher-1 and -forks-1-**

36: (activate-trans philosopher-1 philosopher forks-_-pidp1_11_-rfork state-3 state-4) [1]

37: (queue-write philosopher-0 forks–pid-wfork forks-0- fork) [1]
38: (advance-empty-queue-tail forks-0- queue-1 qs-0 qs-0 fork empty zero one) [1]
39: (perform-trans philosopher-0 philosopher forks–pid-wfork state-1 state-6) [1]
40: (activate-trans philosopher-0 philosopher forks–pid-rfork state-6 state-3) [1]

**41: Macro-Action B, using philosopher-0 and -forks-0-**

44: (activate-trans philosopher-0 philosopher forks-_-pidp1_5_-rfork state-3 state-4) [1]

Figure 16: Plan for the Philosophers problem before macro-action expansion.

The plan shown in Figure 16 was produced by Marvin for problem four in the Philosophers domain (before the translation of the macro-actions back into sequences of single-step actions). The first five steps are found easily through guidance from the heuristic; the following eleven are found during a period of exhaustive search which are, upon exiting the plateau, used to form a macro-action, macro-action A. Macro-action B is formed in a similar manner later in the planning process, and is subsequently used to avoid further exhaustive search. In solution plans for problems involving more philosophers, the two macro-actions are used several times: macro-action A is used once for each consecutive pair of philosophers, and macro-action B once for each odd-numbered philosopher (and once for philosopher-0). The graph in Figure 17 shows the performance of Marvin when the macro-actions are not inferred during search compared to that when the macro-actions are inferred; both configurations produce identical solution plans. It can be seen that the performance is consistently improved when the macro-actions are used, as exhaustive plateau search is avoided.

As can be seen in Figure 18, using macro-actions provides improved performance in the PSR domain: 23 rather than 15 problems can be solved, and in the majority of the cases solved by the two configurations, a solution can be found in less time when macro-actions are used.

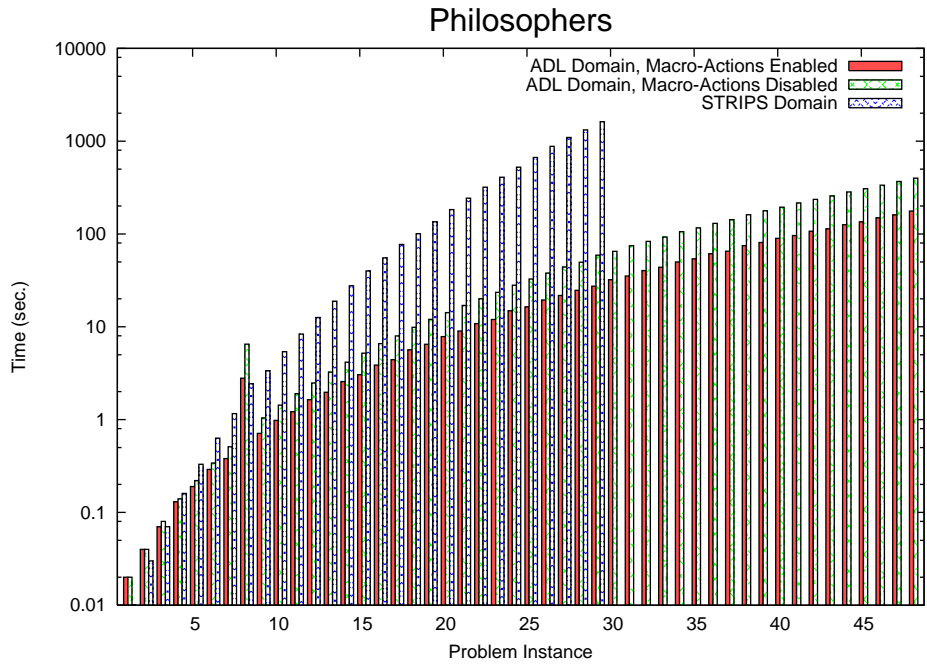

Figure 17: Time taken to find a solution plan in the Philosophers domain with the STRIPS domain encoding and the ADL domain encoding, with and without macro-actions.

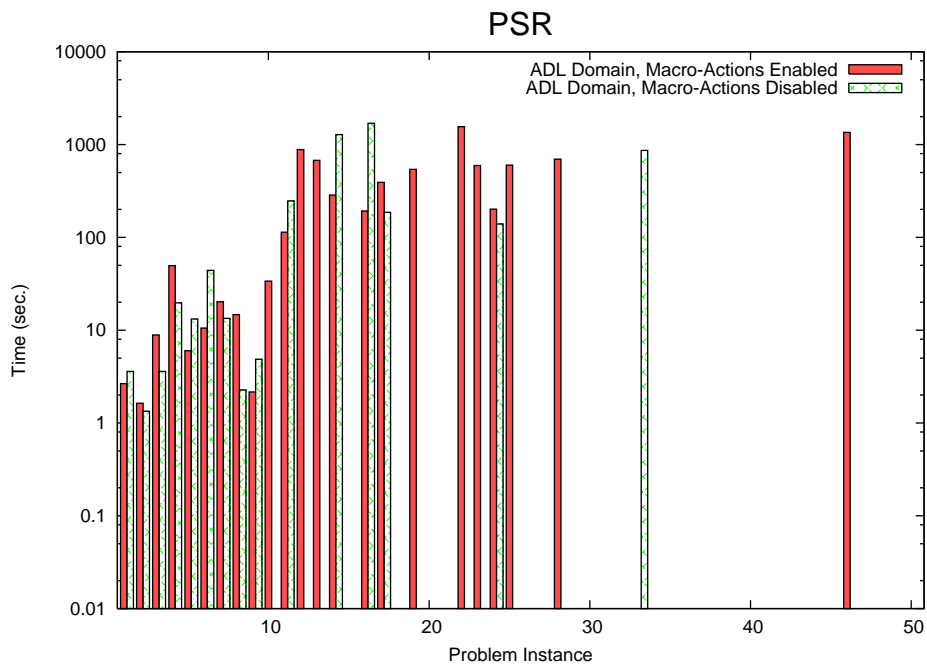

Figure 18: Time taken to solve problems in the PSR domain with and without macro-actions.

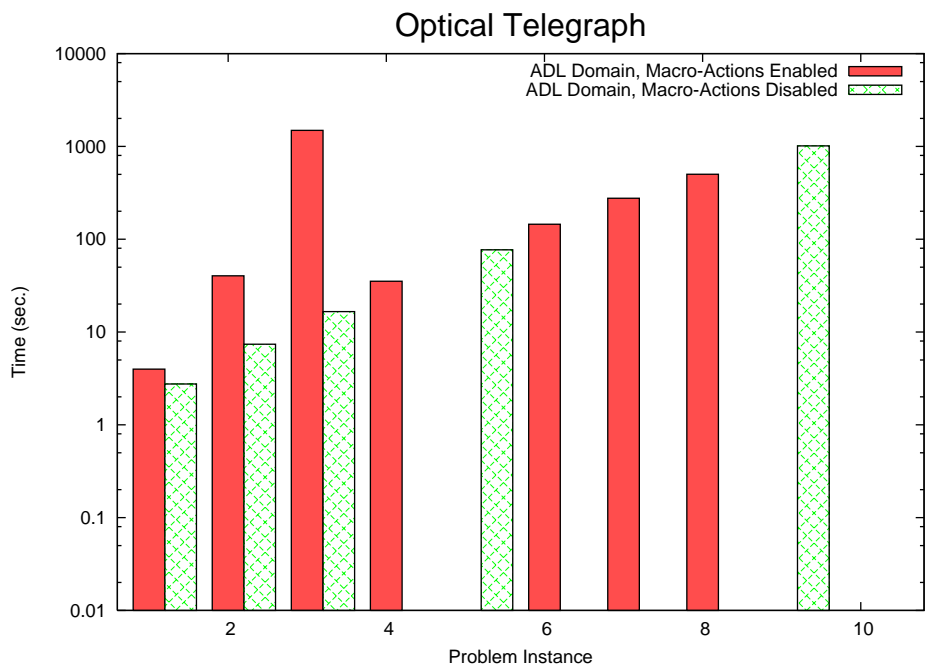

Figure 19: Time taken to solve problems in the Optical Telegraph domain with and without macro-actions.

As shown in Figure 19, Marvin is only able to solve a few of the first 10 problems in the promela/optical-telegraph domain. Nonetheless, when macro-actions are enabled, a net of two additional problems can be solved.

## 7. Future Work

The macro-action strategy adopted by Marvin in IPC 4 was to generate its macro-actions on a *per-problem* basis. It is possible, however, to build libraries of macro-actions on a *per-domain* basis; this approach was taken by Macro-FF (Botea et al., 2005). Marvin's macro-actions could also be cached for use when solving all the problem instances in a given domain. If this were done, then the knowledge encapsulated in the plateau-escaping macro-actions that allows heuristic imperfections in the search landscape to be bypassed could be made available across all the problems in a given domain without needing exhaustive search to re-discover this knowledge on each problem instance. In contrast to existing systems that use off-line learning to generate and test macro-actions, caching Marvin's plateau-escaping macro-actions across solving a problem suite in this manner would allow for online learning to take place. Further work is being undertaken in this area, to investigate effective caching strategies and to manage the large number of macro-actions found.

The idea of using plateau-escaping macro-actions is not restricted to search under the relaxed planning graph heuristic. Currently, the effect of using the macro-actions in search under other heuristics is being investigated, including the causal-graph heuristic (Helmert, 2004) used by Fast-Downward .

At present, the macro-actions used in Marvin are restricted to those used to escape plateaux. Work is currently in progress exploring ways of extending the macro-learning capabilities of Marvin to include more general macro-action structures of the kind being explored by Botea and Schaeffer (Botea et al., 2005).

## 8. Conclusions

We have presented a forward search heuristic planner called Marvin, which introduces several modifications to the search strategy of FF. These are:

- The use of learned macro-actions for escaping plateaux.

- A least-bad-first search strategy for search on plateaux.

- A greedy best-first search strategy when EHC fails.

- The addition of native support for both ADL and derived predicates, without relying on a domain preprocessor.

Results presented indicate that the effects of these modifications varies depending on the domain with which the planner is presented, but can be summarised as:

- The inference and use of plateau-escaping macro-actions:

  - Provides improved performance in the Philosophers, Depots, Driverlog and Pipestankage-nontemporal domains, in terms of planner execution time.

  - Although performance did not improve in the other domains, it did not significantly degrade, with the exception of FreeCell.

  - The makespan of the plans found in the majority of domains was not degraded by the use of macro-actions.

- The use of least-bad-first search:

  - Provides substantial improvements in planner performance in the Philosophers domain.

  - Reduces planner execution time in the Satellite, Driverlog and Pipestankage-nontemporal domains, sometimes at the expense of increased solution plan makespans.

  - Provides worse performance in the FreeCell and Pipesnotankage-nontemporal domains.

- Greedy best-first search does not perform significantly differently from best-first search in the evaluation domains considered.

- Other than in the Airport domain, where no difference in performance is observed, the native support for derived predicates and ADL improves the performance of the planner; either by allowing a more-compact higher-level domain formulation to be used, or by improving the effectiveness of the macro-actions inferred.

## Acknowledgments

We would like to thank the anonymous referees for their comments, and Maria Fox for her help in revising this manuscript. We also thank Derek Long for supporting us in entering Marvin into IPC 4 and Jörg Hoffmann and Stefan Edelkamp for their hard work in organising the competition.